\documentclass[10pt]{article}
\newif\ifblindreview

\usepackage[letterpaper]{geometry}
\usepackage{amta2024}
\usepackage{times}
\usepackage{url}
\usepackage{latexsym}
\usepackage{natbib}
\usepackage{layout}
\usepackage{multicol}
\usepackage{booktabs,array}
\usepackage{float}
\usepackage[hidelinks]{hyperref}
\usepackage[all]{hypcap}
\usepackage{graphicx}
\usepackage{inconsolata}
\usepackage{xcolor}

\usepackage{url}
\usepackage{tabularx}

\usepackage{comment}

\hypersetup{
    colorlinks,
    linkcolor={red!50!black},
    citecolor={blue!50!black},
    urlcolor={blue!80!black}
}

\ifblindreview
  \newcommand{\authorinfo}{\author{}} 
\else
  \newcommand{\authorinfo}{
    \author{\name{\bf Séamus Lankford} \hfill  \addr{seamus.lankford@adaptcentre.ie}\\
            \addr{\small ADAPT Centre, Department of Computer Science, Munster Technological University, Cork, T12P928, Ireland.} \\
            \name{\bf Andy Way} \hfill \addr{andy.way@adaptcentre.ie}\\
            \addr{\small ADAPT Centre, School of Computing, Dublin City University, Dublin, D09DXA0, Ireland.}
    }
  }
\fi

\begin{document}

\amtaHeader{x}{x}{xxx-xxx}{2015}{45-character paper description goes here}{Author(s) initials and last name go here}
\title{\bf Leveraging LLMs for  MT in Crisis Scenarios: \\ a blueprint for low-resource languages}
\authorinfo

\maketitle
\pagestyle{empty}

\begin{abstract}
\vspace{5pt}
In an evolving landscape of crisis communication, the need for robust and adaptable Machine Translation (MT) systems is more pressing than ever, particularly for low-resource languages. This study presents a comprehensive exploration of leveraging Large Language Models (LLMs) and Multilingual LLMs (MLLMs) to enhance MT capabilities in such scenarios. By focusing on the unique challenges posed by crisis situations where speed, accuracy, and the ability to handle a wide range of languages are paramount, this research outlines a novel approach that combines the cutting-edge capabilities of LLMs with fine-tuning techniques and community-driven corpus development strategies. At the core of this study is the development and empirical evaluation of MT systems tailored for two low-resource language pairs, illustrating the process from initial model selection and fine-tuning through to deployment. Bespoke systems are developed and modelled on the recent Covid-19 pandemic. The research highlights the importance of community involvement in creating highly specialised, crisis-specific datasets and compares custom GPTs with NLLB-adapted MLLM models. It identifies fine-tuned MLLM models as offering superior performance compared with their LLM counterparts. A scalable and replicable model for rapid MT system development in crisis scenarios is outlined. Our approach enhances the field of humanitarian technology by offering a blueprint for developing multilingual communication systems during emergencies.
\end{abstract}

\begin{multicols}{2}

\section{Credits}
This work was supported by ADAPT, which is funded under the SFI Research Centres Programme (Grant 13/RC/2016) and is co-funded by the European Regional Development Fund. The research was also funded by the Munster Technological University in Cork, Ireland.

\section{Introduction}
The excitement surrounding LLMs stems from their potential to revolutionise many fields, from language translation~\citep{costa2022no} and content generation~\citep{brown2020language} to chatbots\footnote{https://chatgpt.com} and virtual assistants. \cite{Way2024} observes that from the very outset, machine translation (MT) capability has been overhyped at each paradigm shift over the past 75 years, but with their ability to `understand' language and generate complex responses, LLMs do appear capable of enhancing human communication and productivity in ways that were  unimaginable with previous approaches, especially given that LLMs are not restricted to text-based use-cases, and can be used in creative applications such as generating music\footnote{https://soundraw.io} or art.

When building LLMs, the focus is on designing and training the model architecture. This involves selecting the appropriate neural network architecture and hyperparameters, as well as deciding on the training data and optimisation techniques to use.

Tuning an MLLM or LLM, in contrast, involves adjusting the parameters of a pre-trained model to improve its performance on a specific task. In neural networks such as MLLMs, the weights and biases are parameters that the network adjusts through training to minimise a cost function. This is performed by training the model on a task-specific dataset and adjusting the model's hyperparameters to optimise its performance. Tuning an MLLM can be a challenging task, as the model is often very complex and the training process can take a long time, but \cite{lankford2023adaptMLLM} offer an open-source solution to fine-tuning pre-built MLLMs, with a particular focus on low-resource language pairs, thus overcoming much of this complexity. In contrast to predictions of their imminent demise \citep{vdM2021}, \cite{Way2024} predicts that tools such as adaptMLLM will instead allow translators to gain a competitive edge, by building and tuning their own models with their own high-quality data, ``while retaining full control over the process, leading to self-empowerment and an improved sense of well-being". 

Given their potential, this paper investigates whether tools such as adaptMLLM can be used to rapidly build good-quality MLLM-based MT systems for deployment in crisis scenarios, where speed of development is crucial, but not at the expense of quality altogether. These deployments are contrasted with the development of custom GPTs and fine-tuned LLMs. For two language pairs and four language directions, each featuring a minority language, we present and evaluate a pipeline that we hope can be used as a blueprint for rapid deployment in crisis scenarios to improve multilingual communication.

\section{Background}\label{related}

\cite{way2020} observe that there “have been alarmingly few attempts to provide automatic translation services for use in crisis scenarios”. To the best of our knowledge, the first was Microsoft's effort \citep{lewis2010} to build Haitian Creole systems following the devastating earthquake in 2010, as the title makes clear “from scratch in 4 days, 17 hours, \& 30 minutes”. Estimated casualties ranged from 100,000 to over 300,000 deaths, with around a third of all citizens affected in some way or other by the earthquake measuring 7.0 on the Richter scale. The main issues for the Microsoft team were a complete lack of knowledge of the language (grammatical structure, encoding, orthography etc), and no data at all to train high-quality statistical MT engines. However, the team quickly identified some available resources (the Bible is available in most languages), and a small number of native speakers to help with translation and, especially, validation of the MT output generated. Eventually, around 150,000 segments of training data were collected to build the system, which obtained a BLEU \citep{papineni2002BLEU} score of almost 30 for Creole to English, and 18.3 for English to Creole, sufficiently high (especially for the into-English direction) for the system to be deployed for use by relief workers in the field.

This remarkable effort led to the writing of a cookbook for MT in crisis scenarios \citep{lewis2011}, so that the lessons learned from the exercise could be put into practice when other crises arose, as they do all too commonly, regrettably. Importantly, \cite{lewis2011} note that “If done right, MT can dramatically increase the speed by which relief can be provided”. In any such scenario, translation is almost always needed, and despite its importance, it is often overlooked.

In response to the need for better preparation for translation readiness in crises, Sharon O’Brien coordinated the Interact project\footnote{https://sites.google.com/view/crisistranslation/home} featuring partners from academia, industry, as well as NGOs. \cite{Federici2019} provide a set of recommendations within that project which apply mainly to human translation provision in crisis scenarios. 

\subsection{Multilingual Language Models---NLLB}

MT has become a significant area of research with the aim of eliminating language barriers worldwide. However, the current focus is limited to a small number of languages, neglecting the vast majority of low-resource languages. In an effort to address this issue, the No Language Left Behind (NLLB) initiative was launched to try to overcome the challenges of using MT for low-resource language translation by developing datasets and models that bridge the performance gap between low- and high-resource languages. The NLLB team has also created architectural and training enhancements tailored to support MT for low-resource languages. Their work is open source,\footnote{https://github.com/facebookresearch/fairseq/tree/nllb} and many of their models serve as baselines for fine-tuning with adaptMLLM \citep{lankford2023adaptMLLM}.\footnote{https://github.com/adaptNMT/adaptMLLM} While projects like this are undoubtedly a step in the right direction, \cite{ignat2023} observe that “state-of-the-art MT models such as NLLB-200 … still perform poorly on many low-resource languages, such as African languages” (p.3), so much work remains to be done.

\subsection{Large Language Models}\label{llm_background}

The increasing availability of large datasets provides the raw material for LLM training~\citep{radford2019language,conneau-etal-2020-unsupervised,winata-etal-2021-language},
enabling performance improvement on a wide variety of NLP tasks.  

LLMs have the potential to improve the use of technology across a wide range of domains, including medicine, education and computational linguistics. In education, LLMs may be used for personalised student learning experiences~\citep{KASNECI2023102274}, while in the medical domain, analysing large amounts of medical files can assist doctors in treating patients~\citep{iftikhar2023docgpt}. 
Of particular interest to our research is the manner in which LLMs can be used within the realm of NLP, more specifically in the field of MT, and we now provide details of some of the main candidates in this space.

\subsubsection{GPT-4}

The primary distinction between GPT-3.5 and GPT-4\footnote{https://openai.com/product/gpt-4} is that while the former is a text-to-text model, the latter is more of a data-to-text model, exhibiting the ability to perform tasks that its predecessor could not. For example, GPT-4 is capable of processing visual input as part of a prompt, such as images or web pages, and can even generate text that explains the humour in memes. Consequently, GPT-4 can be classified as a ``multimodal model''. Furthermore, GPT-4 has a longer memory than its previous versions, with a short-term memory closer to 64,000 words, enabling it to maintain coherence during extended interactions. GPT-4 also enables users to select different personalities for the model's responses.

OpenAI has not disclosed the number of parameters used in the training of GPT-4, though many estimates suggest it may be around 1.76 trillion. However, other sources, such as AX Semantics,\footnote{https://en.ax-semantics.com} have estimated the number to be around 100 trillion, with such a large model costing around \$100 million to build. AX Semantics maintains that such a number makes the language model (LM) more akin to the functioning of the human brain with respect to language and logic.
  
\subsubsection{Gemini}

Gemini\footnote{https://gemini.google.com} comes in three versions tailored for varying levels of complexity and application: Gemini Ultra for the most demanding tasks, Gemini Pro for a broad range of activities, and Gemini Nano for on-device applications. The Ultra variant, in particular, has demonstrated SOTA performance, outperforming human benchmarks in massive multitask language understanding (MMLU) across a suite of 57 subjects. \citet{Gemini2024} documents the performance of Gemini on the “Machine Translation from One Book (MTOB)” benchmark \citep{Tanzeretal2023}, essentially how good a model is at learning a language from almost no resources. For an evaluation of Gemini 1.5 Pro on the FLORES-200 benchmark \citep{costa2022no} against Google Translate, GPT-3.5 and GPT-4, and other systems, see \cite{Akteretal2023} (p.12).

\subsubsection{CoPilot}
Microsoft has introduced Microsoft 365 Copilot,\footnote{https://copilot.microsoft.com} a generative AI tool designed to enhance workplace productivity and creativity. Copilot integrates LLMs with user data from Microsoft Graph and Microsoft 365 apps, to allow users to utilise natural language commands across familiar Microsoft 365 applications such as Word, Excel and PowerPoint.

Central to this announcement is the launch of Business Chat, which synergies with the LLM, Microsoft 365 apps, and user data to generate outputs such as status updates from natural language prompts, drawing from various data sources like emails, meetings, and documents. This ensures that users remain in control, enabling them to adjust or refine the outputs as needed.

\section{Datasets}\label{datasets}

\subsection{Language Pairs}
To benchmark the translation performance of adaptMLLM in fine-tuning MLLMs for low-resource languages, we had to choose suitable language pairs for which appropriate datasets existed. The English-to-Irish (EN${\leftrightarrow}$GA) and English-to-Marathi (EN${\leftrightarrow}$MR) language pairs were selected since they fulfilled the criteria of low-resource languages, and data was freely available from shared tasks featuring these language pairs in crisis scenarios. Therefore, these language pairs were very suitable for evaluating our proposed pipeline for rapidly generating high-quality translations in crisis situations by fine-tuning MLLMs. 

 Irish is the first official language of the Republic of Ireland, and is also recognised as a minority language in Northern Ireland. Irish is an official language of the European Union and a recognised minority language in Northern Ireland with an ISO code of ``GA''.\footnote{https://www.iso.org}

The dominant language spoken in India's Maharashtra state is Marathi, with an ISO code of ``MR''. It has over 83 million speakers, and it is a member of the Indo-Aryan language family. Despite being spoken by a significant number of people, Marathi is considered to be relatively under-resourced when compared to other languages used \mbox{in the region.}

\subsection{Shared Task Datasets}

To benchmark the performance of our adaptMLLM-trained models, datasets from the LoResMT2021 shared task \citep{ojha2021findings} were used, since the shared task focused on low-resource languages  including both  EN${\leftrightarrow}$GA  and  EN${\leftrightarrow}$MR in the specific domain of translation of COVID-related data. 

The datasets from the shared task provided 502 Irish and 500 Marathi validation sentences whereas 250 (GA${\rightarrow}$EN), 500 (EN${\rightarrow}$GA), and 500 (EN${\leftrightarrow}$MR) sentences were made available in the test datasets. Training data consisted of 20,933 lines of parallel data for the EN${\leftrightarrow}$MR language pair and 13,171 lines of parallel data were used to train the EN${\leftrightarrow}$GA models. A detailed breakdown of all resources is available in \cite{ojha2021findings}.

\section{Approach}\label{approach}

\begin{figure}[H]
    \centering
    {\includegraphics[width=\linewidth]{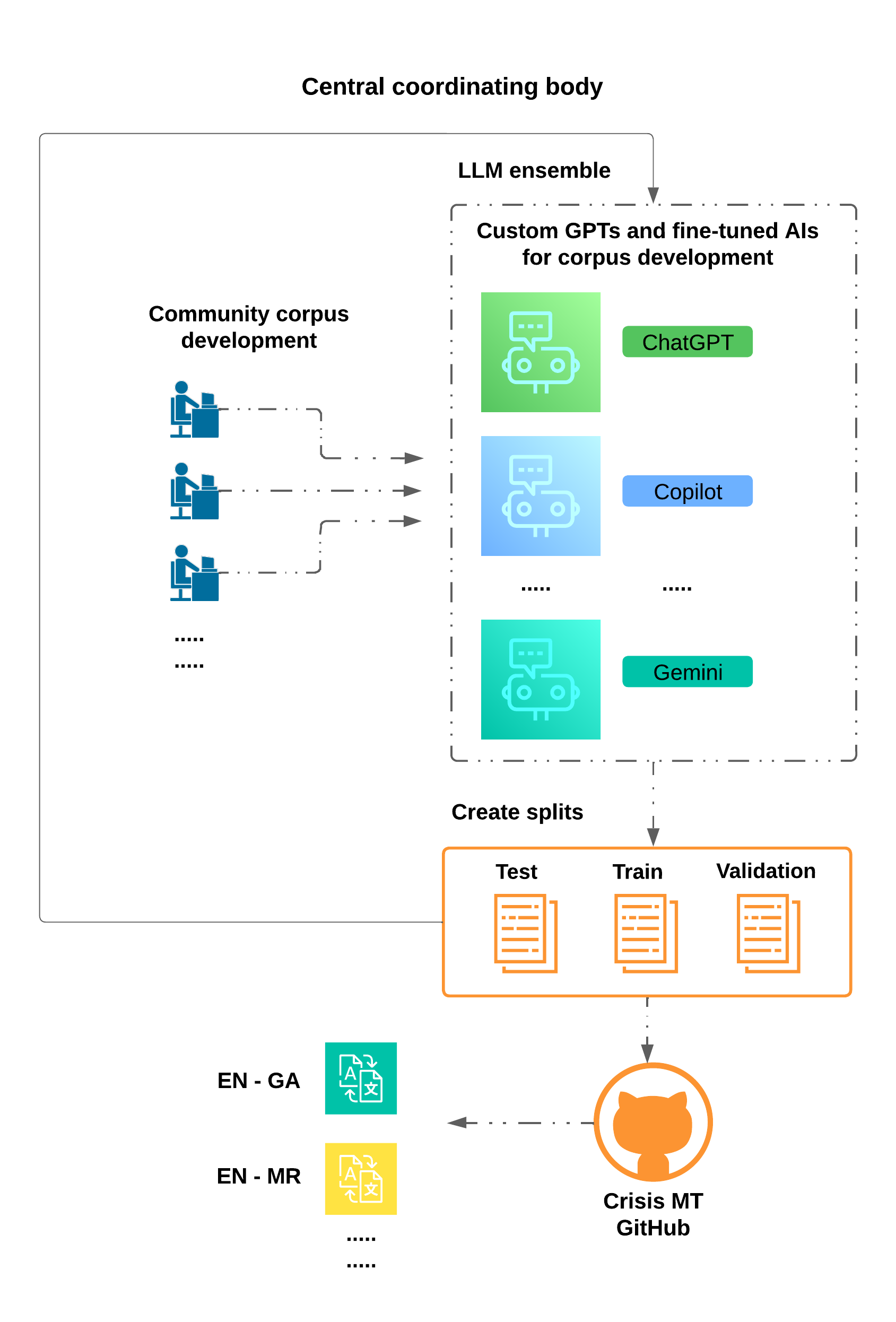}}
    \caption{Community corpus development using custom GPTs from a range of foundation models.}
    \label{fig:ensemble_llm}
\end{figure}

Our approach to enhancing MT in crisis situations involves three key elements. Initially, a custom GPT would be created on the ChatGPT platform immediately after a crisis, enabling users to contribute to a specialised knowledge base with new terms relevant to the crisis, effectively crowd sourcing a dataset for crisis-specific language pairs. With this approach both in-domain corpora and simple first iteration models are developed in real time by disparate users entering source and reference translations. Within the custom GPT interface on ChatGPT the functionality also exists to upload relevant documents which adds to the knowledge base of custom GPTs. Another interesting feature of ChatGPT is its ability to publicly share custom GPTs by sharing links. In this manner, it is trivial to develop corpora by implementing a simple link-sharing strategy that invites community-wide, expert-only or an ensemble of contributions.  

As the crisis evolves, these corpora are then used to develop more accurate MT models with new weights tailored to the specific language needs of the crisis by fine-tuning OpenAI models, or other LLM foundation models. 

Finally, a bespoke model could be created using an open-source tool like adaptMLLM, fine-tuned with a custom dataset developed during the crisis. Such a phased approach allows for a rapid initial response and progressively more tailored MT solutions as the crisis unfolds, leveraging community input and specialised training to improve translation accuracy in critical situations.

Of course, a major consideration when designing an MT system in crisis scenarios is the availability of suitable parallel corpora which contain new terminology associated with the unfolding crisis. However, it is precisely at these times when the production of such datasets presents the greatest challenge. 

Figure \ref{fig:ensemble_llm} presents a structured approach to developing language corpora with community involvement, using customised LMs, and preparing the data for MT projects which are shared on GitHub. A central coordinating body (such as ACL,\footnote{https://aclanthology.org} AMTA,\footnote{https://amtaweb.org} EAMT\footnote{https://eamt.org} or an equally invested stakeholder) could oversee the process working in conjunction with relevant industry partners and other stakeholders.
 
There are two parallel streams in this process, the first of which entails a community corpus development effort, involving multiple contributors, using a collaborative, crowdsourced approach. In this phase, selected users and language experts interact with LLMs on an {\it ad hoc} basis by presenting text in the source language and providing the translation in the target language. In this manner, an in-domain parallel dataset relevant specifically to the particular crisis is rapidly developed for the chosen language pair.

The second stream, LLM ensemble, incorporates several elements: models from ChatGPT, Copilot, Gemini and other foundation models. The corpus creation process is carried out by simply exporting and concatenating the conversation histories from each of the customised LLMs. Duplicate entries created in the corpus development stage are removed and the corpus is split into three datasets: ``Test'', ``Train'', and ``Validation''. The training dataset is used to fine-tune a pre-trained (M)LLM  to create a bespoke in-domain crisis MT model. The validation dataset is also used as part of this fine-tuning process before the test set is used to evaluate the performance of the MT system using standard BLEU, TER~\citep{snover2006study} and ChrF \citep{popovic2015ChrF} metrics.

Finally, the output of the process feeds into a crisis MT GitHub which is the central repository for the development of MT systems for multiple language pairs. Models and datasets developed as part of this process would be shared on GitHub for open-source collaboration and distribution.

A Colab notebook has been developed to help with this process and we have made it publicly available as part of this paper's GitHub which is freely available for download.\footnote{https://github.com/adaptNMT/crisisMT/blob/main/communitycorpus.ipynb} A Gradio-based web app is incorporated within this notebook which facilitates the involvement of non-technical users in corpus creation. This is our first implementation of such a notebook for aiding crisis MT corpus development and as an open-source tool, improvements and contributions from the community are welcomed.

\begin{figure}[H]
    \centering
    {\includegraphics[width=\linewidth]{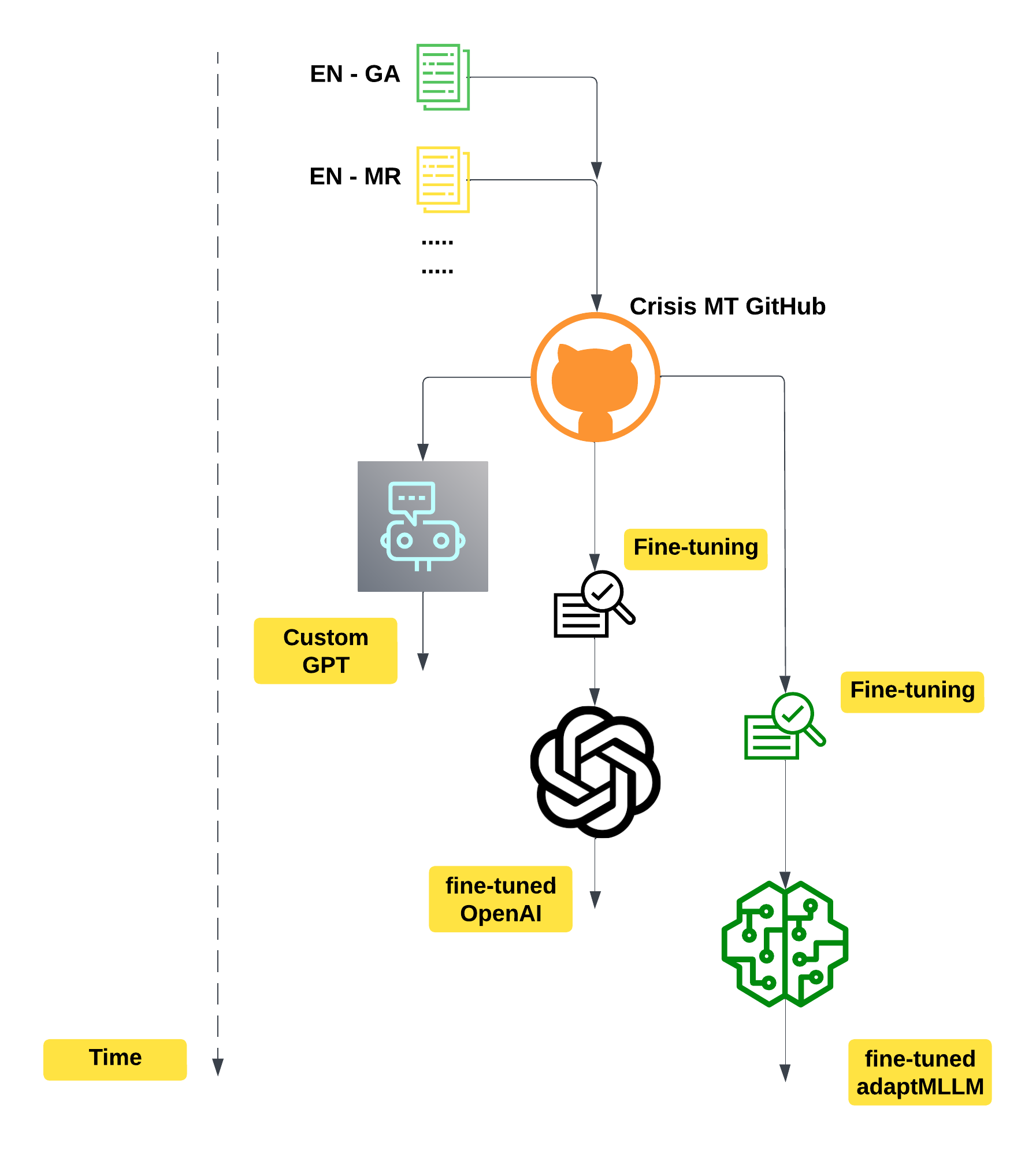}}
    \caption{Model development process.}
    \label{fig:model_dev}
\end{figure}

The elements of model development required to manage MT in a crisis are highlighted in Figure \ref{fig:model_dev}. All models use the outputs from the corpus development process. In the initial phase of the crisis, a custom GPT is created from the crowd-sourced corpus development. At this point, a parallel track is in progress where a fine-tuned LLM, such as a pre-trained OpenAI model, is developed and made available at a later date. Our approach proposes a third parallel track which develops a fine-tuned MLLM-specific model using a tool such as adaptMLLM. Fine-tuning an open-source MLLM using adaptMLLM has been shown to empirically deliver the highest translation performance (see Section \ref{experiments}). Subsequent phases of model development would also benefit from the availability of a significantly expanded crisis-relevant corpus via the ongoing crowd-sourcing effort. The links to the custom GPTs developed for both language pairs have been open-sourced.\footnotemark{}

\section{Empirical Evaluation}\label{experiments}

\label{sec:exp}
After outlining the details of our approach, the quality of the models developed is evaluated by training models for the EN${\leftrightarrow}$GA and the EN${\leftrightarrow}$MR language pairs.

\subsection{Infrastructure and Hyperparameters}

All MLLM models were trained by fine-tuning a 3.3B parameter NLLB pre-trained model using the adaptMLLM application with a Google Colab Pro+ subscription. The DeepSpeed library enables our models to be loaded across both GPU and system memory, thus reducing the required compute resources. The optimal hyperparameters used for developing models for both language pairs are the same as those identified by \citet{lankford2023adaptMLLM}.

Both the custom GPT models and the baseline models used the GPT-4 model under a standard ChatGPT subscription. The OpenAI fine-tuned models were developed using a pay-as-you-go plan. In fine-tuning the OpenAI models, GPT-3.5-turbo-0125 was the chosen pre-trained model since GPT-4 was unavailable for fine-tuning. Default parameters were kept and the number of epochs was set to auto. For inference on these models, a temperature setting of 0.5 was chosen to ensure a more deterministic output which aligns with the requirements for translation models.

\subsection{Results: Automatic Evaluation}
To determine the quality of our translations, automated metrics were employed. For comparison with  previous results, the performance of our new models was measured using three automatic evaluation metrics: BLEU, TER, and ChrF. We report case-insensitive BLEU scores at the corpus level. Note that BLEU and ChrF are precision-based metrics, so higher scores are better, whereas TER is an error-based metric so lower scores indicate better translation quality. All models, notebooks and translations generated as part of our experiments are freely available for download.\footnotemark[\value{footnote}] \footnotetext{https://github.com/adaptNMT/crisisMT}

\subsubsection{Translation in the EN$\leftrightarrow$GA Directions}
The experimental results in the EN$\leftrightarrow$GA directions are summarised in Tables \ref{en2ga} - \ref{ga2en} and are compared with the baseline highest scores from the LoResMT2021 Shared Task.\footnote{https://machinetranslate.org/loresmt-2021}

The highest-performing EN$\rightarrow$GA system in the LoResMT2021 Shared Task was submitted by ADAPT~\citep{lankford2021machine}. The model was developed with an in-house application, adaptNMT \citep{lankford2023adaptNMT} using a Transformer \citep{Vaswani2017} architecture. It performed well across all key translation metrics (BLEU: 36.0, TER: 0.531 and ChrF3: 0.6).

\begin{table*}
\centering
\begin{tabular}{ll}
\hline
\textbf{Hyperparameter} & \textbf{Values} \\ \hline
Epochs & 1, 3, \textbf{5} \\ \hline
Batch size & 8, 12, \textbf{16} \\ \hline
Gradient steps & 2,  4, \textbf{8} \\ \hline
Learning rate & 1$\times10^{-5}$, \textbf{3$\times10^{-5}$}, 9$\times10^{-5}$ \\ \hline
Weight decay & 0.01, \textbf{0.1}, 1, 2 \\ \hline
Mixed precision & False, \textbf{True} \\ \hline
\end{tabular}
\caption{HPO with optimal hyperparameters highlighted in bold}
\label{hpotable}
\end{table*}

By fine-tuning the NLLB MLLM, using the parameters outlined in Table \ref{hpotable}, a significant improvement in translation performance was achieved. The adaptMLLM EN$\rightarrow$GA system, shown in Table \ref{en2ga}, achieves a BLEU score of 41.2, which is 5.2 BLEU points higher (14\% relative improvement) than the  score of the winning system in 2021. 

Both the custom GPT-4 and GPT-4 baseline models performed well compared to the GPT-3 models. However, there was a significant differential when compared to the adaptMLLM fine-tuned NLLB models which recorded an increase of 8.4 BLEU points which corresponds to a relative improvement of 25\%. In a crisis scenario, a GPT-4 baseline model would be available in real-time. A custom GPT could be available within a matter of minutes once a relevant training corpus is uploaded to the GPT's knowledge base. Such approaches would be suitable for assisting translators in the immediate aftermath of a crisis and would help in issuing bilingual press releases. However, more detailed documentation would greatly benefit from the improved translation quality of a bespoke fine-tuned adaptMLLM solution.

\begin{table*}[!htb]

    \begin{minipage}{.5\linewidth}
      
      \begin{tabular}{lllll}
\hline
\textbf{System}&\textbf{BLEU} & \textbf{TER} & \textbf{ChrF3} \\ \hline
adaptMLLM & 41.2 & 0.51 & 0.48 \\
adaptNMT & 36.0 & 0.531 & 0.60 \\
custom GPT-4 & 32.8 & 0.553 & 0.594 \\
GPT-4 baseline & 31.1 & 0.564 & 0.584 \\
adaptMLLM-base & 29.7 & 0.595 & 0.559 \\
fine-tuned GPT-3.5 & 22.7 & 0.701 & 0.488 \\
GPT-3.5 baseline & 20.0 & 0.712 & 0.475 \\
\hline
\end{tabular}
\caption{EN$\rightarrow$GA}
\label{en2ga}
    \end{minipage}%
    \begin{minipage}{.5\linewidth}
      \centering
        
\begin{tabular}{lllll}
\hline
\textbf{System}& \textbf{BLEU}&\textbf{TER}&\textbf{ChrF3} \\ \hline
adaptMLLM & 75.1 & 0.385 & 0.71 \\
GPT-4 baseline & 53.9 & 0.365 & 0.754 \\
custom GPT-4 & 53.2 & 0.37 & 0.747 \\
fine-tuned GPT-3.5 & 50.2 & 0.419 & 0.713 \\
adaptMLLM-base & 47.8 & 0.442 & 0.692 \\
GPT-3.5 baseline & 41.6 & 0.512 & 0.668 \\
IIITT & 34.6 & 0.586 & 0.61\\ 
\hline
\end{tabular}
\caption{GA$\rightarrow$EN}
\label{ga2en}
    \end{minipage} 
\end{table*}

For translation in the GA$\rightarrow$EN direction, illustrated in Table \ref{ga2en}, the best-performing model for the LoResMT2021 Shared Task was developed by IIITT with a BLEU score of 34.6, a TER score of 0.586 and ChrF3 score of 0.6. Accordingly, this serves as the baseline score by which we can benchmark our GA$\rightarrow$EN MLLM model, developed by fine-tuning a 3.3B parameter NLLB using adaptMLLM. Similar to the results achieved in the EN$\rightarrow$GA direction, significant improvement in translation performance was observed using this new method. The performance of the adaptMLLM model offers an improvement across all metrics with a BLEU score of 75.1, a TER of 0.385 and a ChrF3 result of 0.71. In particular, the 117\% relative improvement in BLEU score against the IIITT system is very significant. 
 
The results from our GA$\rightarrow$EN experiments reinforce the findings derived from translating in the EN$\rightarrow$GA direction. The custom and baseline GPT-4 models  immediately deliver a translation system with good quality BLEU scores of 53 points. However, a higher-quality translation system with a 21.2 BLEU score improvement can delivered in a matter of hours once a fine-tuned adaptMLLM NLLB sytem is put in place. The exact length of time for system development is dependent on the quality of the underlying training infrastructure and also, more importantly, on how rapidly the training corpus can be assembled.   

\subsubsection{Translation in the EN$\leftrightarrow$MR Directions}

The experimental results from the LoResMT2021 Shared Task in the EN$\leftrightarrow$MR directions are summarised in Tables \ref{en2mr} and \ref{mr2en}, and are compared with adaptMLLM. For the shared task, the highest-performing EN$\rightarrow$MR system was submitted by the IIITT team. Their model used a Transformer architecture and achieved a BLEU score of 24.2, a TER of 0.59, and ChrF3 of 0.597. 

\begin{table*}[!htb]

    \begin{minipage}{.5\linewidth}
      
      \begin{tabular}{lllll}
\hline
\textbf{System}&\textbf{BLEU} & \textbf{TER} & \textbf{ChrF3} \\ \hline
adaptMLLM & 26.4 & 0.56 & 0.608 \\
IIITT & 24.2 & 0.59 & 0.597 \\
adaptMLLM-base & 19.8 & 0.656 & 0.57 \\ 
custom GPT-4 & 19.0 & 0.678 & 0.528 \\
GPT-4 baseline & 18.5 & 0.689 & 0.527 \\
fine-tuned GPT-3.5 & 9.9 & 0.894 & 0.442 \\
GPT-3.5 baseline & 6.7 & 1.06 & 0.392 \\

\hline
\end{tabular}
\caption{EN$\rightarrow$MR}
\label{en2mr}
    \end{minipage}%
    \begin{minipage}{.5\linewidth}
      \centering
        
\begin{tabular}{lllll}
\hline
\textbf{System}& \textbf{BLEU}&\textbf{TER}&\textbf{ChrF3} \\ \hline
adaptMLLM & 52.6 & 0.409 & 0.704 \\
adaptMLLM-base & 42.7 & 0.506 & 0.639 \\
custom GPT-4  & 38.8 & 0.539 & 0.626 \\
GPT-4 baseline  & 38.6 & 0.546 & 0.617 \\
oneNLP-IIITH  & 31.3 & 0.58 & 0.646 \\
GPT-3.5 baseline & 27.9 & 0.688 & 0.568 \\
fine-tuned GPT-3.5 & 27.6 & 0.716 & 0.501 \\
\hline
\end{tabular}
\caption{MR$\rightarrow$EN}
\label{mr2en}
    \end{minipage} 
\end{table*}

Again the approach taken by adaptMLLM in fine-tuning a 3.3.B parameter NLLB MLLM yielded the best performance  compared with other systems entered for the shared task. The EN$\rightarrow$MR adaptMLLM system achieves the highest BLEU score of 26.4, a 2.2 point improvement (9\% relative) compared with IIITT, the winning team in the EN$\rightarrow$MR shared task.  

The MLLM-based system, trained using adaptMLLM, is also compared with GPT-4 and GPT-3.5 LLM-based systems. For the purposes of our experiments, the best-performing LLM used a custom GPT-4 model which recorded a BLEU score of 19.0 points in the EN$\rightarrow$MR direction. This was only a marginal improvement on the baseline GPT-4 model with a BLEU score of 18.5 points. Critically, however, this solution could be delivered in real time which makes such a model a potential starting point for an immediate crisis response. A relative improvement of 42\% and 7.9 BLEU points is available once sufficient time is given to developing the fine-tuned MLLM model.

\par
For translation in the MR$\rightarrow$EN direction, the best-performing model for the LoResMT2021 Shared Task was developed by oneNLP-IIITT with a BLEU score of 31.3, a TER of 0.58 and ChrF3 of 0.646. This serves as the baseline against which our MR$\rightarrow$EN model, developed using adaptMLLM, can be benchmarked. The performance of the adaptMLLM model offers a significant improvement across all metrics with a BLEU score of 52.6, a TER of 0.409 and a ChrF3 of 0.704. Again this represents a very strong relative improvement of 68\% in BLEU compared with the winning team from the shared task.

The best-performing MLLM-based system in the MR$\rightarrow$EN direction is also compared with  our LLM-based systems. The highest-performing LLM used a custom GPT-4 model which recorded a BLEU score of 38.8 points. This was only a marginal improvement on the baseline GPT-4 model with a BLEU score of 38.6 points. As previously noted, the GPT4 baseline solutions can be delivered in real time which makes this model the ideal starting point for an immediate crisis response. A relative improvement of 36\% and 14 BLEU points is available once sufficient time is given to developing the fine-tuned MLLM model.

\section{Discussion}\label{discussion}
A significant finding of this research is the demonstrated capability to substantially improve translation quality for low-resource languages through fine-tuning with crisis-specific datasets. The adaptability and speed of deployment offered by LLMs and MLLMs hold the promise of making such rapid response a standard practice in future crises, ensuring that linguistic barriers do not impede vital aid and information flow.

However, this potential comes with its share of challenges, particularly concerning the assembly and quality of training datasets. This study's proposed solution, leveraging community input through custom GPTs to crowd-source and refine translation data, presents a scalable model for corpus development in crisis scenarios. Looking ahead, this research lays the groundwork for expanding the application of LLMs and MLLMs beyond MT to address a wider range of NLP challenges in crisis situations. The blueprint provided for rapid MT system deployment in emergencies, emphasising community involvement and model fine-tuning, offers valuable insights for future endeavours aiming to harness AI for humanitarian purposes.

\section{Conclusion}\label{concl}
In this paper, we outlined how the advent of LLMs has transformed our ability to rapidly develop MT systems for low-resource languages in crisis scenarios. A system for rapid corpus development was presented which adopts a collaborative approach, emphasising community involvement and open-source methodologies. 

The appropriate response to developing MT systems at different phases of a crisis were highlighted. Using the recent Covid pandemic as a reference crisis, MT systems were developed using custom GPTs, fine-tuned models from OpenAI and fine-tuned MLLM models. We demonstrated that a custom GPT delivers a functioning MT system rapidly whereas a fine-tuned MLLM delivers a higher-quality solution given a longer time horizon.    

By highlighting how a fine-tuned MLLM can provide SOTA accuracy during a crisis, our work demonstrates how LLMs and MLLMs can provide more inclusive communication. Language barriers in crisis communication will be diminished with the help of this approach which in turn helps minority communities in times of real need.

Our paper introduces a pipeline which is applicable to a broader range of NLP problems. As part of future work, the methodologies and insights derived from our research could extend beyond the scope of MT to other domains within NLP. Consequently, a versatile framework for addressing a variety of language processing challenges in crisis scenarios has been put forth in this study.

\section*{Limitations of study}\label{limitations}

The proprietary nature of MLLMs and LLMs such as NLLB and GPT-4, which do not disclose the specifics of their training datasets presents a problem. When fine-tuning these models for specific tasks, there is a risk of overlapping data that cannot be easily identified or removed. This limitation underscores a broader issue within the field of NLP and MT research, where the exact composition of training data in SOTA models often remains opaque.



\begin{small}
\bibliographystyle{apalike}
\bibliography{amta2024}

\begin{thebibliography}{}

\bibitem[Akter et~al., 2023]{Akteretal2023}
Akter, S.~N., Yu, Z., Muhamed, A., Ou, T., Bäuerle, A., Cabrera, A.~A., Dholakia, K., Xiong, C., and Neubig, G. (2023).
\newblock An in-depth look at gemini's language abilities.
\newblock eprint arXiv:2312.11444.

\bibitem[Brown et~al., 2020]{brown2020language}
Brown, T., Mann, B., Ryder, N., Subbiah, M., Kaplan, J., Dhariwal, P., Neelakantan, A., Shyam, P., Sastry, G., and Askell, A. (2020).
\newblock Language models are few-shot learners.
\newblock In {\em Proceedings of the 34th Conference on Neural Information Processing Systems (NeurIPS 2020)}, volume~33, pages 1877--1901, Vancouver, BC, Canada.

\bibitem[Conneau et~al., 2020]{conneau-etal-2020-unsupervised}
Conneau, A., Khandelwal, K., Goyal, N., Chaudhary, V., Wenzek, G., Guzmán, F., Grave, E., Ott, M., Zettlemoyer, L., and Stoyanov, V. (2020).
\newblock Unsupervised cross-lingual representation learning at scale.
\newblock In {\em Proceedings of the 58th Annual Meeting of the Association for Computational Linguistics}, pages 8440--8451, Online.

\bibitem[Costa-jussà et~al., 2022]{costa2022no}
Costa-jussà, M., Cross, J., Çelebi, O., Elbayad, M., Heafield, K., Heffernan, K., Kalbassi, E., Lam, J., Licht, D., and Maillard, J. (2022).
\newblock No language left behind: Scaling human-centered machine translation.
\newblock arXiv.

\bibitem[Federici et~al., 2019]{Federici2019}
Federici, F., O’Brien, S., Cadwell, P., Marlowe, J., Gerber, B., and Davis, O. (2019).
\newblock International network in crisis translation—recommendations on policies.
\newblock 11p.

\bibitem[Iftikhar et~al., 2023]{iftikhar2023docgpt}
Iftikhar, L., Iftikhar, M., and Hanif, M. (2023).
\newblock Docgpt: Impact of chatgpt-3 on health services as a virtual doctor.
\newblock {\em EC Paediatr.}, 12:45--55.

\bibitem[Ignat et~al., 2023]{ignat2023}
Ignat, O., Jin, Z., Abzaliev, A., Biester, L., Castro, S., Deng, N., Gao, X., Gunal, A., He, J., Kazemi, A., Khalifa, M., Koh, N., Lee, A., Liu, S., Min, D., Mori, S., Nwatu, J., Perez-Rosas, V., Shen, S., Wang, Z., Wu, W., and Mihalcea, R. (2023).
\newblock A phd student's perspective on research in nlp in the era of very large language models.
\newblock eprint arXiv:2305.12544.

\bibitem[Kasneci et~al., 2023]{KASNECI2023102274}
Kasneci, E., Sessler, K., Küchemann, S., Bannert, M., Dementieva, D., Fischer, F., Gasser, U., Groh, G., Günnemann, S., and Hüllermeier, E. (2023).
\newblock Chatgpt for good? on opportunities and challenges of large language models for education.
\newblock {\em Learn. Individ. Differ.}, 103:102274.

\bibitem[Lankford et~al., 2021]{lankford2021machine}
Lankford, S., Afli, H., and Way, A. (2021).
\newblock Machine translation in the covid domain: An english-irish case study for loresmt 2021.
\newblock In {\em Proceedings of the 4th Workshop on Technologies for MT of Low Resource Languages (LoResMT2021)}, pages 144--150, Virtual.

\bibitem[Lankford et~al., 2023a]{lankford2023adaptMLLM}
Lankford, S., Afli, H., and Way, A. (2023a).
\newblock adaptmllm: Fine-tuning multilingual language models on low-resource languages with integrated llm playgrounds.
\newblock {\em Information}, 14:638.

\bibitem[Lankford et~al., 2023b]{lankford2023adaptNMT}
Lankford, S., Afli, H., and Way, A. (2023b).
\newblock adaptnmt: An open-source, language-agnostic development environment for neural machine translation.
\newblock {\em Lang. Resour. Eval.}, 57:1671--1696.

\bibitem[Lewis, 2010]{lewis2010}
Lewis, W. (2010).
\newblock Haitian creole: How to build and ship an mt engine from scratch in 4 days, 17 hours, \& 30 minutes.
\newblock In {\em EAMT-2010: Proceedings of the 14th Annual Conference of the European Association for Machine Translation}, St. Raphael, France.
\newblock 6p.

\bibitem[Lewis et~al., 2011]{lewis2011}
Lewis, W., Munro, R., and Vogel, S. (2011).
\newblock Crisis mt: Developing a cookbook for mt in crisis situations.
\newblock In {\em Proceedings of the Sixth Workshop on Statistical Machine Translation}, page 501–511.

\bibitem[Ojha et~al., 2021]{ojha2021findings}
Ojha, A., Liu, C., Kann, K., Ortega, J., Shatam, S., and Fransen, T. (2021).
\newblock Findings of the loresmt 2021 shared task on covid and sign language for low-resource languages.
\newblock In {\em Proceedings of the 4th Workshop on Technologies for MT of Low Resource Languages (LoResMT2021)}, pages 114--123, Virtual.

\bibitem[Papineni et~al., 2002]{papineni2002BLEU}
Papineni, K., Roukos, S., Ward, T., and Zhu, W. (2002).
\newblock Bleu: A method for automatic evaluation of machine translation.
\newblock In {\em Proceedings of the 40th annual meeting of the Association for Computational Linguistics}, pages 311--318, Philadelphia, PA, USA.

\bibitem[Popovi{\'c}, 2015]{popovic2015ChrF}
Popovi{\'c}, M. (2015).
\newblock chrf: character n-gram f-score for automatic mt evaluation.
\newblock In {\em Proceedings of the Tenth Workshop on Statistical Machine Translation}, pages 392--395, Lisboa, Portugal.

\bibitem[Radford et~al., 2019]{radford2019language}
Radford, A., Wu, J., Child, R., Luan, D., Amodei, D., and Sutskever, I. (2019).
\newblock Language models are unsupervised multitask learners.
\newblock OpenAI Blog.

\bibitem[Snover et~al., 2006]{snover2006study}
Snover, M., Dorr, B., Schwartz, R., Micciulla, L., and Makhoul, J. (2006).
\newblock A study of translation edit rate with targeted human annotation.
\newblock In {\em Proceedings of the 7th Conference of the Association for Machine Translation in the Americas: Technical Papers}, Cambridge, MA, USA. Citeseer.
\newblock Forest Grove, OR, USA.

\bibitem[Tanzer et~al., 2023]{Tanzeretal2023}
Tanzer, G., Suzgun, M., Visser, E., Jurafsky, D., and Melas-Kyriazi, L. (2023).
\newblock A benchmark for learning to translate a new language from one grammar book.
\newblock eprint arXiv:2309.16575.

\bibitem[Team, 2024]{Gemini2024}
Team, G. (2024).
\newblock Gemini 1.5: Unlocking multimodal understanding across millions of tokens of context.
\newblock Google blog.

\bibitem[van~der Meer, 2021]{vdM2021}
van~der Meer, J. (2021).
\newblock Translation economics of the 2020s: A journey into the future of the translation industry in eight episodes.
\newblock {\em Multilingual Magazine}.
\newblock July/Aug 2021.

\bibitem[Vaswani et~al., 2017]{Vaswani2017}
Vaswani, A., Shazeer, N., Parmar, N., Uszkoreit, J., Jones, L., Gomez, A., Kaiser, L., and Polosukhin, I. (2017).
\newblock Attention is all you need.
\newblock In {\em Proceedings of the 31st International Conference on Neural Information Processing Systems}, pages 6000--6010, Long Beach, CA, USA.

\bibitem[Way, 2024]{Way2024}
Way, A. (2024).
\newblock {\em What does the Future hold for Translation Technologies in Society?}
\newblock Routledge, Abingdon, Oxon, UK.
\newblock In S. Baumgarten and M. Tieber (eds.) Routledge Handbook of Translation Technology and Society, forthcoming.

\bibitem[Way et~al., 2020]{way2020}
Way, A., Haque, R., Xie, G., Gaspari, F., Popovic, M., and Poncelas, P. (2020).
\newblock Rapid development of competitive translation engines for access to multilingual covid-19 information.
\newblock {\em Informatics}, 7(2):21.

\bibitem[Winata et~al., 2021]{winata-etal-2021-language}
Winata, G., Madotto, A., Lin, Z., Liu, R., Yosinski, J., and Fung, P. (2021).
\newblock Language models are few-shot multilingual learners.
\newblock In {\em Proceedings of the 1st Workshop on Multilingual Representation Learning}, pages 1--15, Punta Cana, Dominican Republic.

\end{thebibliography}
\end{small}

\end{multicols}

\end{document}